# Style Obfuscation by Invariance

**Chris Emmery**
Tilburg University
University of Antwerp
cmry@protonmail.com

**Enrique Manjavacas**
University of Antwerp
enrique.manjavacas
@uantwerpen.be

**Grzegorz Chrupała**
Tilburg University
g.chrupala@uvt.nl

## Abstract

The task of obfuscating writing style using sequence models has previously been investigated under the framework of obfuscation-by-transfer, where the input text is explicitly rewritten in another style. These approaches also often lead to major alterations to the semantic content of the input. In this work, we propose obfuscation-by-invariance, and investigate to what extent models trained to be explicitly style-invariant preserve semantics. We evaluate our architectures on parallel and non-parallel corpora, and compare automatic and human evaluations on the obfuscated sentences. Our experiments show that style classifier performance can be reduced to chance level, whilst the automatic evaluation of the output is seemingly equal to models applying style-transfer. However, based on human evaluation we demonstrate a trade-off between the level of obfuscation and the observed quality of the output in terms of meaning preservation and grammaticality.

## 1 Introduction

The fact that writing style uniquely characterizes a person, and can be leveraged for automatic author identification (Holmes, 1998; Stamatatos et al., 2000), has been well-studied in the field of (computational) stylometry (Neal et al., 2017). Similarly, work on author profiling (Koppel et al., 2002) has demonstrated that such stylometric features can be used to accurately infer an extensive set of personal information, such as age, gender, education, socio-economic status, and mental health issues (Eisenstein et al., 2011; Alowibdi et al., 2013; Volkova et al., 2014; Plank and Hovy, 2015; Volkova and Bachrach, 2016). Traditionally, these techniques relied on expensive human-labelled examples; however, more recent work has demonstrated near equal accuracy when only relying on self-reports as a distant supervision signal (Beller et al., 2014; Emmery et al., 2017; Yates et al., 2017). While these efforts have been greatly beneficial to various research fields such as computational sociolinguistics (Daelemans, 2013), the resulting techniques potentially expose users of such media to directed attacks where this information can be abused unbeknownst to them. This is particularly harmful to individuals in a vulnerable position regarding race, political affiliation, mental health, or any other personal information that they made explicitly unavailable.

Adversarial stylometry, or style obfuscation, is one of the proposed methods aimed at protecting users against such attacks. Its objective is to rewrite an input text such that a classifier (the adversary) trained on detecting a particular variable (such as a demographic attribute) is fooled — effectively protecting this variable. The main challenge is to preserve the original meaning of the input, whilst hiding the act of obfuscation (Potthast et al., 2016). Recent work on automatic obfuscation (Shetty et al., 2017; Karadzhov et al., 2017) shows promising results in minimizing performance of the adversary; however, these models (and as noted by the authors) struggle with correctly maintaining the semantic content of the input. To illustrate, while rewriting *school* to *wedding*[1] effectively fools an age classifier into thinking the text is written by an adult rather than a teen, the original meaning is not preserved in the output.

---

[1]Example taken from Shetty et al. (2017).

We propose that this observed shift in meaning is to some extent a by-product of the formulation of the obfuscation task. Content words that are strongly related to a particular attribute often play a significant role in the accuracy of a potential adversary. There is ample evidence for this phenomenon in age and gender classification work (Koppel et al., 2002; Rao et al., 2010; Burger et al., 2011; Sap et al., 2014, inter alia). Taking examples from Sap et al. (2014) specifically, features with strong coefficient weights for gender include e.g. *boxers, shaved, girlfriend, beard, fightin* for males, and *purse, blueberry, pedicure, hubby, earrings* for females. It is therefore not a surprising result that models explicitly tasked to *minimize* the performance of such classifiers perform what we will refer to as obfuscation-by-transfer. To illustrate, the adversary is easily fooled when a sentence looks strongly female even though it was written by a male. As such, the easiest route to obfuscation from this perspective is a form of style-transfer: swapping a few strongly target-associated content words for their contrastive variant (*wife* to *husband*, *school* to *wedding*). When such variants are also close in semantic spaces that sequence models make use of, any reconstruction metrics—such as BLEU (Papineni et al., 2002), METEOR (Banerjee and Lavie, 2005), embedding distances, etc.—might become an inaccurate indication of the change in meaning.

**Our contributions** In this work we propose a different approach to automatic obfuscation that we hypothesize partly overcomes the limitations to preserving meaning of the input: obfuscation-by-invariance. Here, the objective shifts towards maximizing adversary's *uncertainty*, implying its accuracy on the protected variable should be as close to chance level as possible. Fixing the adversary's performance around chance involves making the input text devoid of stylistic features that strongly correlate with any of the protected variables, thus producing language that is neutral with respect to these style differences. We test our hypothesis in several experimental conditions.[2] The main component in our encoder-decoder architecture to achieve a style-invariant encoding of the input is a Gradient Reversal Layer (GRL) (Ganin and Lempitsky, 2015) inserted between the input sentence embedding and the style classifier. We investigate the effect of this module in isolation, as well as in a style-invariant soft transfer setting by using a conditioned decoder (Ficler and Goldberg, 2017). First, to gauge if this architecture can successfully decode from the style-invariant encoding—and what the effect on adversary performance would be—we train sequence-to-sequence models on a parallel corpus of English Bible styles. Secondly, given that in a realistic obfuscation setting there is no access to such parallel sources, we drop the target pairs to create an autoencoder setting. In our experiments, we demonstrate a trade-off around chance-level performance: obfuscation-by-transfer in a parallel setting works well using a many-to-many translation model, but scores worse in the human evaluation than our style-invariant model. As such, we pose that there is potential in an style-invariant approach to obfuscation, and it deserves further investigation.

## 2 Related Work

### 2.1 Adversarial Stylometry

The idea that computational stylometry might be used to compromise anonymity was first explored by Rao et al. (2000). They saw potential to conceal style information in machine translation (MT), but noted that it was not powerful enough at the time. Kacmarcik and Gamon (2006) continued the proposed line of work by informing users regarding characteristic features and deeper linguistic cues in their writing style. Recent related studies can be found in (Caliskan-Islam et al., 2015; Le et al., 2015). Brennan et al. (2012) explicitly frame obfuscation as an adversarial task and use MT (round-trip translation), similar to (Caliskan and Greenstadt, 2012). Rule-based perturbations (Juola and Vescovi, 2011) and mixtures of both (Karadzhov et al., 2017) have also been applied for fully automatic obfuscation. Closest to our approach is recent work by Shetty et al. (2017), who pursue the task of learning obfuscation-by-transfer using a Generative Adversarial Network architecture. In their setup, a generator is trained to produce sentences that maximize the probability assigned by a discriminator that is, in turn, trained to distinguish real from generated sentences. While they incorporate different semantic losses, and demonstrate successful obfuscation on age and gender annotated micro-blog data and political speeches, their output suffers from the lack of semantic preservation we described before. Finally, a style-transfer approach with potential application to obfuscation is work by Carlson et al. (2017), who investigate textual zero-shot style-transfer using a sequence-to-sequence MT model inspired by zero-shot translation (Johnson et al., 2016). Their work demonstrates successful translation between many different versions of the English Bible.

---

[2] All code and data to fully replicate the experiments will be made available soon.

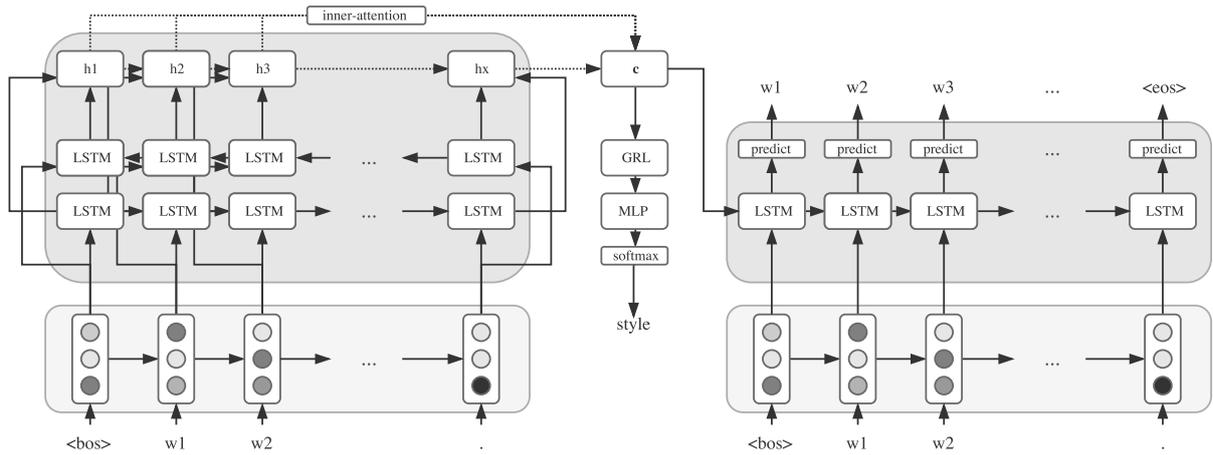

Figure 1: Model architecture with the Gradient Reversal Layer (GRL) module: Left the encoder part, middle the GRL taking the input sentence embedding (also called context vector) **c** as input, and right the decoder part of the architecture. Note that via the GRL working on the context vector, the encoder part of the model tries to minimize the style classification performance of the MLP, whilst still having to produce a useful representation for the decoder.

## 2.2 Gradient Reversal

The use of a Gradient Reversal Layer (GRL) for learning domain invariant feature representations was proposed by Ganin and Lempitsky (2015), who demonstrated its viability for learning lightning-condition invariance in computer vision. Since then, it has been applied to several language tasks: e.g. textual feature extraction (Pryzant et al., 2017), POS tagging (Kim et al., 2017; Gui et al., 2017), image captioning (Chen et al., 2017), and document classification (Liu et al., 2017; Xu and Yang, 2017). Most importantly, Xie et al. (2017) demonstrate the GRL module can be used to implement an adversarial setting, and to improve performance for a number of language tasks, including generation. These results bode well for its application to obfuscation-by-invariance.

## 3 Models

Our base architecture is a neural encoder-decoder (Sutskever et al., 2014) model similar to that of (Wu et al., 2016), implemented in PyTorch (Paszke et al., 2017). Given an input sequence of one-hot encoded words, the encoder first embeds the words into dense vectors which are then processed by one or more bidirectional (Schuster and Paliwal, 1997) layers of Long Short-Term Memory (Hochreiter and Schmidhuber, 1997) (LSTM) cells. The resulting sequence of processed vectors is then merged into a single dense representation using an inner-attention mechanism that will be described below. After encoding, a neural language-model is trained to decode the output sequence conditioning on the sentence embedding (so called context vector) resulting from the encoder. Training is accomplished by minimizing the per-word cross-entropy loss of the target sequence.

In an autoencoder setting, the goal of the network is to simply reconstruct the original sentence based on the encoded context vector (Lauly et al., ; Li et al., 2015). This set-up can be combined with the GRL to encourage the encoder to produce attribute-invariant context vectors. The target is the input itself in the case of an autoencoder (AE) architecture, or a paired sentence in the case of a sequence-to-sequence

(S2S) architecture. See Figure 1 for a visual representations of the base architecture.

### 3.1 Architecture Components

In addition to the architecture described above, we introduce a few extra components:

**Gradient Reversal Layer (GRL)** The GRL (Sutskever et al., 2014) is applied on top of the encoder output. During the forward pass the GRL is the identity function which feeds input to a shallow Multi-Layer Perceptron (MLP) style classifier. However, during back-propagation the gradient of the loss is flipped in sign. The idea is that the encoder parameters are optimized to generate sentence embeddings that do not contain information which can be used to recover the style of the input.

**Conditional Decoder (C)** Previous research on neural language modelling has shown the effectiveness of conditioning a language model on sentential and contextual variables (such as tense, modality or voice). In our experiments, we evaluate a conditional autoencoder in which the decoder is conditioned on the input style label. We implement conditioning following the approach by Ficler et al. (2017), which simply concatenates the corresponding attribute embedding vector (in our case, the corresponding style embedding) to each of the word embeddings input to the decoder. Each style is therefore associated with an embedding vector **c**, which is fed into the architecture at each step. In contrast to the simple autoencoder, the conditional autoencoder allows to choose a desired style at test time. We suspect that by encouraging the decoder to target a certain style, the output would be more consistent without fully recovering the targeted style.

**Token Transfer (TT)** It can be argued that an MT system relying on a parallel corpus of styles (be it attributes or authors) would perform obfuscation-by-transfer. Moreover, it would likely largely preserve the original meaning as translation is a meaning-preserving operation (Ide et al., 2002; Dyvik, 2004). However, such parallel corpora are generally not available and have very high associated compilation costs, as it would require large amounts of identical information (ideally on sentence-level) to be written by e.g. teens and adults. Textual style transfer by MT is therefore not a plausible use-case for obfuscation. However, it does provide a good indication of the performance of an obfuscation model under the framework of obfuscation-by-transfer. For this, we apply a sequence-to-sequence translation model trained on style as discussed by (Carlson et al., 2017). Following the work of (Johnson et al., 2016), we use a target style token, allowing for a model trained on many-to-many translation that can be used to rewrite an input sentence in a different style simply by prepending a target token. This is the only configuration that uses Bahdanau attention (Bahdanau et al., 2014), as it is not tested in combination with the GRL (which tends to negate the style token).

### 3.2 Architecture Details

All our models use 300-dimensional embeddings. Both the encoder and the decoder are implemented with a single layer of 1000 LSTM cells. The sequence of hidden states from the encoder are merged into a single representation using a feature-wise inner-attention mechanism. Let $w_t$ and $h_t$ denote respectively the input word embedding and hidden state of the LSTM for step $t$ for a total of $n$ total steps. The $i^{th}$ feature of the sentence embedding $c$ is computed by a weighted sum over the $i^{th}$ feature of the hidden states:

$$c_i = \sum_{t=1}^{n} a_i^t h_i^t \quad (1)$$

where each weight $a_i^t$ is computed by:

$$a_i^t = \frac{\exp([W^T z^t]_i)}{\sum_{s=1}^{n} \exp([W^T z^s]_i)} \quad (2)$$

In Equation 2, $z_t$ stays for the concatenation of $w_t$ and $h_t$, $W \in \mathbb{R}^{(D+H) \times H}$ is an additional projection matrix to be learned, and $D$ and $H$ denote the dimensionalities of the word embedding matrix and the LSTM layer respectively. In comparison to traditional merging models (such as max or mean pooling), the additional parameters help the model to learn what input words and what features are more important for the task. This is similar to conventional attention over hidden activation vectors with the main difference being that weighting is done feature-wise and all information flow from the encoder to the decoder is passed through the bottleneck of the single output encoding vector. Note that a traditional alignment-style attention mechanism is not compatible with the application of the GRL, since the latter must have scope over all information being passed from the encoder to the decoder.

The decoder is conditioned by the output encoding by concatenating the encoding vector to the input of the decoder at each step. This facilitates learning by increasing the gradient signal to the encoder. All model parameters are optimized with Adam (Kingma and Ba, 2014) in mini-batches of 50 examples and a learning rate of 0.001 which is decreased by 0.75 after each epoch. To avoid overfitting, dropout (Srivastava et al., 2014) is applied to the input of each LSTM layer with a dropping probability of 0.25 and we stop training when loss stops decreasing for 3 epochs. During test time, we approximate the model best output sequence with beam search and a beam of size 5. GRL is implemented with a single-layer MLP with dimensionality matching the one of the encoder.

## 4 Experimental Set-up

Our main goal is investigating the effectiveness of obfuscation-by-invariance, and more specifically to what extent style-invariant representations preserve sentential semantics of the input. To this end, we perform three experiments for different corpora and obfuscation settings, using the components described above. In each of these settings, there is an adversary trained to detect the to be obfuscated variable. We describe our experiments and evaluations below.

### 4.1 Data

We use a highly parallel corpus of five different versions of the English bible (retrieved from GitHub[3], originally collected from `openbible.info`[4]). These versions are semantically consistent, but vary stylistically across different aspects; BBE is written in simplified English, YLT follows the syntactic structures of the original Greek and Hebrew, and other versions (KJV, DBY, ASV) correspond to older editions reflecting diachronic variations. The sentence-level verse coding (book + chapter + verse ID) is used for almost perfect pairing between the different versions (some missing pairs were removed), forming style quintuplets, which were tokenized using `spaCy`[5] (Honnibal and Montani, 2017). The style tuples are divided amongst train (80%), dev (10%), and test (10%) sets—ensuring all sentences in the development and test splits are unseen, regardless of the style combination that they occur in—and all style combinations (excluding a same-style combination) are used to form 620,752 pairs.

### 4.2 Adversary

The adversary is a sentence-level classifier in the form of `fastText`[6]; a simple linear model with one hidden embedding layer that learns sentence representations using bag of words or $n$-gram input, producing a probability distribution over the given styles using the softmax function. The classifier is trained on the source side of the training split, as these are the instances we intend to obfuscate. It is run for 20 epochs using 100 dimensional embeddings, uni and bi-grams, a learning rate of 0.01, and a bucket size of 1M. It achieves an accuracy of 86.6%, and chance level performance for the adversary is 20% given 5 classes with an even distribution.

### 4.3 Evaluation

To automatically evaluate the reconstruction and semantic preservation of our generated sentences, we use MT metrics, as well as distance in semantic embedding space. Obfuscation is measured by the difference of the adversary's accuracy compared to chance level.

---

[3] https://github.com/scrollmapper/bible_databases
[4] http://www.openbible.info/labs/cross-references/
[5] https://spacy.io/
[6] https://github.com/facebookresearch/fastText

**MT Metrics (BLEU, METEOR)**  We calculate BLEU-4 and METEOR (Papineni et al., 2002; Banerjee and Lavie, 2005) using `nlg-eval`[7]. Given that this is not a standard MT task, we provide these scores between the generated sentence and the source sentence ($\leftarrow$), as well as the generated sentence and the target sentence ($\rightarrow$). For the sequence-to-sequence models, $\rightarrow$ is the primary indicator of successful obfuscation-by-transfer. However, $\leftarrow$ gives some indication how much the output is still related to the original. Introducing the GRL to this models should decrease scores for both $\leftarrow$ and $\rightarrow$. For the autoencoder, we only have $\leftarrow$ to evaluate, which should generally strongly decrease when adding the GRL.

**Word Mover's Distance (WMD)**  To assess the word embedding distance of the obfuscated sentence to the original, we take the Word Mover's Distance (WMD) (Kusner et al., 2015), based on the English `fastText` embeddings for Wikipedia (Bojanowski et al., 2016). WMD takes the distance between two sentences in a weighted point cloud of embedded words as the minimum cumulative distance that this cloud for sentence $A$ needs to travel to align with that of sentence $B$, and is shown to capture some semantic relations.

**Adversary Impact ($\Delta$ acc)**  We compare the accuracy of the adversary on the generated sentence to that of the original to asses obfuscation strength. However, as our goal is to keep the adversary's performance level close to chance, we define $\Delta$ accuracy $=$ accuracy $- p$ where $p$ is majority baseline. Therefore, if $\Delta$ accuracy is negative, this means the adversary's performance has dropped below chance, and the task is closer to obfuscation-by-transfer rather than by-invariance. Subsequently, a positive score indicates the extent to which obfuscation fails to match both cases.

**Gaussian Noise ($\mu$)**  To enforce a significant change in the decoded output, one can simply add a Gaussian noise mask to the context vector during generation time. We generate this mask as a random vector from a Gaussian Distribution $N(0, \mu)$, where $\mu = \{0.01, 0.05, 0.10, 0.15, 0.20\}$ and add this to the values of the context vector. This noise can be utilized to increase obfuscation (due to more random decoding behaviour) at the cost of quality of the output.

### 4.4 Experiments

Using all components discussed above, we define three experimental settings to measure the effect of applying a GRL and conditional decoder to achieve obfuscation-by-invariance. (**1**) We train our architecture on the style pairs from the English Bible corpus in a many-to-many sequence-to-sequence setting. By introducing a GRL here, words that are highly indicative of the target style are not captured by the encoder. To achieve an effective many-to-many MT system (and thus style-transfer) setting we prepend the `<2{stylename}>` token. (**2**) We train an autoencoder on disconnected pairs. Here we introduce both the GRL, plus the conditioned decoder. We hypothesize that the conditioned decoder allows for soft style-transfer from a neutral encoding, implying it would preserve semantic structure better than the MT model. (**3**) We use Gaussian noise to make the sequence-to-sequence and autoencoder models equivalent in obfuscation performance to allow for direct comparison.

## 5 Results

### 5.1 Experimental Results

All results and automatic evaluations are shown in Table 1. As we hypothesized, using style-transfer for obfuscation works well overall, performing either at, or below chance level. However, looking at the target BLEU and METEOR, the sequence-to-sequence model without the target token generates sentences that are closer to source than they are to the target; and achieves low scores overall, with the sentences being quite far off based on WMD. However, note that this is many-to-many translation without any signal regarding the target, given languages with largely the same vocabulary. As such TT is a more realistic reflection of style-transfer success. In terms of translation quality it barely improves over the original baseline, but it does successfully perform obfuscation-by-transfer, as indicated by the 12% accuracy below chance. Assessing the performance of the GRL in this setting, it does not seem improve translation, nor obfuscation — which was largely in line with our expectations.

---
[7] https://github.com/Maluuba/nlg-eval

| model | C | GRL | TT | PPL | BLEU ← | MTR ← | BLEU → | MTR → | WMD | Δ ACC |
|---|---|---|---|---|---|---|---|---|---|---|
| s2s | | | | 9.08 | 22.0 | 25.3 | 17.8 | 23.6 | 1.50 | 1.6 |
| s2s | | x | | 9.27 | 21.8 | 25.2 | 16.9 | 22.9 | 1.50 | 6.9 |
| s2s | | | x | 7.38 | 34.9 | 30.5 | 39.2 | 29.9 | 1.24 | -12.0 |
| AE | | | | 1.51 | 79.5 | 52.9 | - | - | 0.25 | 64.4 |
| AE | | x | | 1.99 | 60.0 | 41.2 | - | - | 0.65 | 50.8 |
| AE | x | x | | 1.87 | 51.9 | 38.1 | - | - | 0.79 | 18.3 |
| source | | | | | 100.0 | 100.0 | 36.0 | 34.5 | 0.00 | 66.6 |

Table 1: Results for the Bible experiments. The first column (model) indicates the setting of our base architecture: either sequence-to-sequence (S2S), or autoencoder (AE). The second column specifies which modules were incorporated: C for the conditioned decoder, GRL for the Gradient Reversal Layer, and TT for the prepended style token. The results show perplexity on the dev set (PPL), BLEU and METEOR between source (←) / target (→) and the obfuscated sentence, the Word Mover's Distance score between source and the obfuscated sentence (WMD) and the extent to which the obfuscated sentence pushes the adversary to chance level performance (Δ ACC). In the last row, we note a 'source' baseline, that is achieved by simply copying the source. This shows how overall how much source and target overlap, and the actual above-chance performance of the adversary.

| | AE | | | AE + GRL | | | AE + C + GRL | | |
|---|---|---|---|---|---|---|---|---|---|
| μ | BLEU ← | MTR ← | Δ ACC | BLEU ← | MTR ← | Δ ACC | BLEU ← | MTR ← | Δ ACC |
| - | 79.5 | 52.9 | 64.4 | 60.0 | 41.2 | 50.8 | 51.9 | 38.1 | 18.3 |
| 0.01 | 78.7 | 52.2 | 64.1 | 59.7 | 41.4 | 50.5 | 52.0 | 38.5 | 18.5 |
| 0.05 | 52.9 | 38.9 | 57.2 | 54.8 | 39.1 | 47.6 | 49.1 | 37.2 | 16.5 |
| 0.10 | 14.7 | 21.6 | 36.1 | 32.8 | 32.7 | 36.9 | 40.7 | 33.6 | 11.6 |
| 0.15 | 4.2 | 14.7 | 23.3 | 25.3 | 26.2 | 24.3 | 30.0 | 29.1 | 4.6 |
| 0.20 | 1.5 | 11.3 | 16.4 | 15.3 | 21.5 | 15.5 | 21.1 | 24.9 | 0.0 |

Table 2: Effect of adding Gaussian Noise ($\mu$) to the autoencoder.

The autoencoder setting provides a clearer look into the performance of the GRL, in particular in terms of obfuscation-by-invariance. The plain autoencoder to some extent successfully reproduces the target; looking at BLEU, the adversary performance, and WMD, it is still closely related to the input. Introducing the GRL does impact the relation to the source sentence, but does gain little in comparison on obfuscation performance. However, when the conditioned decoder is added to the architecture, obfuscation performance is visibly impacted more than the decrease in BLEU and METEOR. Lastly, the effect of adding a Gaussian noise mask on the decoder can be found in Table 2. Around $0.15$, the metrics seem to be largely comparable in terms of BLEU, METEOR and Δ accuracy. We further investigate the three most suitable models (S2S + TT, AE + GRL + C, and AE + GRL + C + $\mu = 0.15$) in a human evaluation.

### 5.2 Human Evaluation

For the human evaluation, 20 pairs (original, obfuscated) were sampled from the output of the three models of interest (making 60 pairs in total). Each pair was rated by four participants (all with a linguistics background), on three dimensions using a three point scale. These dimensions included semantic consistency between the original and the obfuscated sentence, the syntactical coherence of the latter, and the amount of changes in the output. The participants were made aware of which sentence of the pairs was the original, and were explicitly asked to rate the dimensions with the original as reference. The participants were not aware that there were multiple models, and the pairs were shuffled so that compar-ing between pairs with the same original was impossible. To simplify the comparison to the original, we only sampled from BBE (basic English Bible). See Table 4 for a sample of the instructions given to the participants, and Table 3 for the results.

|   | SEMANTICS | GRAMMATICALITY | CHANGES |
|---|---|---|---|
| 1 | Semantics are broken; sentence does not mean the same. | Part(s) of, or the complete sentence is garbled. | Change in special characters or flipping a single word. |
| 2 | Slight semantic change, but not intrusive. | Slight word order change that is ungrammatical. | Multiple words were changed, but they align with the original. |
| 3 | Semantics are intact, changes do not alter the meaning. | Grammaticality has not been affected. | New parts have been introduced / rewritten in the sentence. |

Table 3: Explanations per rating for the three dimensions in the human evaluation study.

|   | SEMANTICS | GRAMMATICALITY | CHANGES |
|---|---|---|---|
| S2S + TT | 1.88 | 2.21 | 2.43 |
| AE + GRL + C | 2.12 | 2.32 | 1.99 |
| AE + GRL + C + $\mu = 0.15$ | 1.35 | 1.66 | 2.65 |

Table 4: Average scores and standard deviation for the three dimensions in the human evaluation.

According to the evaluation results, AE + GRL + C has the overall preference of the raters in all three dimensions. Specifically, given that we are interested in semantic preservation, this model is evaluated better than a style-transfer model that has some access to semantic relations between source and target on the SEMANTICS dimension. Note that based on the CHANGES dimension, the AE + GRL + C is the most conservative, which is in line with the BLEU and METEOR scores in Table 1, and likely propagates into the GRAMMATICALITY dimension.

### 5.3 Qualitative Analysis

In this section we perform a manual comparison of the text generated by the three models that were evaluated by our raters in the previous section. Accordingly, we will identify the strengths and weaknesses of our experimental approach, and propose possible lines of future work.

One of the issues we identified with obfuscation-by-transfer was that of small, localized changes in the input, specifically focussing on words that are most relevant to the adversary. When looking at longer sentences such as Table 5, some (semi-)correct variants can be found in e.g. town → city, waste land → dry land, and in Table 6, beryl → onyx, stamp → seal, but incorrect ones also remain rest → work. Overall, the longer the sequence, the more variation can be observed.

A more interesting observation is that some parts of sentences are added to by the models, e.g. *there is without a waste land ; and she makes an dry land*—while incorrectly inserting '*without*', the rest can be considered is a correct expansion. The same holds for Table 6, where *fixed in twisted frames of gold* is expanded to *whereupon they bound in the skillfully woven red frames of gold* , partly erroneously, similar to *living men* in Table 5. Contrastively, the autoencoder + GRL in particular also seems to favour somewhat compressed phrases, removing adjectives such as *young* in *young ox*, *strong* in *strong city*, *beryl* in *beryl stones* and not incorporating *an unpeopled living* altogether.

When directly comparing the sequence-to-sequence and autoencoder + GRL examples, it can be inferred that the transfer approach seems (at least in these examples) quite conservative, sticking to an almost exact alignment, and only making small changes. This however also causes the autoencoder to replace words with unrelated variants or insert not directly related ones; the same behaviour can also be observed in the sequence-to-sequence, however.

The output of the autoencoder shows some evidence that actual *rewrites* of the sentence are possible, which is potentially an interesting path to pursue. Including different variables in the conditioned decoder would make experiments in this direction feasible. It must be noted that evidence of style-neutral rewrites is difficult to find in the Bible; not only due to the archaic constructions, but more so due to the fact that it requires a level of expertise to recognize style shifts. Applying the autoencoder to data that is non-parallel at least has some ground given the current results, and should definitely be part of future work.

| ORIGINAL | For the strong town is without men, an unpeopled living - place; and she has become a waste land: there the young ox will take his rest, and its branches will be food for him. |
|---|---|
| S2S + TT | For the strong city is powerless, an astonishment living and she is become a corrupt land: a young ox shall rest, and its branches shall be for him.' |
| AE + GRL + C | For the city is without living men, there is without a waste land; and she makes an dry land: an ox shall take his work, and their branches shall be food for him. |
| AE + GRL + C + $\mu = 0.15$ | A man dwelleth without an dry land; and, wandering she - place; men shall there become a waste a land: an ox - goat shall take his horses, and his branches shall be prepare for him . |

Table 5: Example 1 — Isaiah 27:10

| ORIGINAL | Then they made the beryl stones, fixed in twisted frames of gold and cut like the cutting of a stamp, with the names of the children of Israel.' |
|---|---|
| S2S + TT | Then they made the onyx stones as a hundred stones, burning in engraved stones of gold, and cut as the marks of a seal, with the names of the children of Israel. |
| AE + GRL + C | Then they wrote the stones, whereupon they bound in the skillfully woven red frames of gold and made like the cutting of a stamp, with the names of the children of Israel.' |
| AE + GRL + C + $\mu = 0.15$ | Then they presented the pillars that belonged in Henadad. The bottom of fine gold and made like the jewels of a stamp of them, at the dial of the children of Israel.' |

Table 6: Example 2 — Exodus 39:6

## 6 Conclusion

We presented an alternative framing of the task of automatic style obfuscation—obfuscation-by-invariance—and tested several components in a neural encoder-decoder architecture that were hypothesized to achieve style-invariant rewrites of the input text. We tested the effect of a Gradient Reversal Layer and a conditioned decoder for obfuscation in parallel and non-parallel settings. Although strong evidence for style-neutral text was difficult to find for the Bible corpus, we demonstrated through human evaluation that our autoencoder architecture trained on non-parallel data obtained a better evaluation than a model trained on parallel data with partial access to semantic relations between source and target. In our qualitative analysis we found evidence for semantically correct local changes of the input, as well as partial rewrites that fit the context of the verses. These results bode well for extending this architecture to other non-parallel corpora to test obfuscation in a practical use-case, e.g. author attributes such as age and gender.